\documentclass[letterpaper, 10 pt, conference]{ieeeconf}  

\IEEEoverridecommandlockouts                              

\usepackage{graphicx}
\usepackage{subfigure}
\usepackage{tabularx}
\usepackage{dblfloatfix}
\usepackage{textcomp}
\usepackage{hyperref}
\usepackage{color}
\usepackage[numbers]{natbib}
\usepackage{siunitx}

\definecolor{NavyBlue}{rgb}{0,0,0.6}

\usepackage{graphics} 
\usepackage{amsmath} 
\usepackage{amssymb}  

\title{\LARGE \bf
Active Clothing Material Perception using Tactile Sensing \\ and Deep Learning
}

\author{Wenzhen Yuan$^{1}$, Yuchen Mo$^{1,2}$, Shaoxiong Wang$^{1}$, and Edward H. Adelson$^{1}$
	\thanks{$^{1}$Computer Science and Artificial Intelligence Laboratory (CSAIL), MIT, Cambridge, MA 02139, USA}
	\thanks{$^{2}$ Department of Computer Science and Technology, Tsinghua University, Beijing, 100084, China }
}

\begin{document}

\maketitle
\thispagestyle{empty}
\pagestyle{empty}

\begin{abstract}
Humans represent and discriminate the objects in the same category using their properties, and an intelligent robot should be able to do the same. In this paper, we build a robot system that can autonomously perceive the object properties through touch. We work on the common object category of clothing.
The robot moves under the guidance of an external Kinect sensor, and squeezes the clothes with a GelSight tactile sensor, then it recognizes the 11 properties of the clothing according to the tactile data. 
Those properties include the physical properties, like thickness, fuzziness, softness and durability, and semantic properties, like wearing season and preferred washing methods. 
We collect a dataset of 153 varied pieces of clothes, and conduct 6616 robot exploring iterations on them. 
To extract the useful information from the high-dimensional sensory output, we applied Convolutional Neural Networks (CNN) on the tactile data for recognizing the clothing properties, and on the Kinect depth images for selecting exploration locations. 
Experiments show that using the trained neural networks, the robot can autonomously explore the unknown clothes and learn their properties. 
This work proposes a new framework for active tactile perception system with vision-touch system, and has potential to enable robots to help humans with varied clothing related housework.


\end{abstract}

\section{Introduction}
A core requirement for intelligent robots is to understand the physical world, which contains understanding the properties of physical objects in the real-world environment. Among the common objects, clothing is an important part. Humans evaluate an article of clothes largely according to its material properties, such as thick or thin, fuzzy or smooth, stretchable or not, etc. The understanding of the clothes' properties helps us to better manage, maintain and wash the clothes. If a robot is to assist humans in daily life, understanding those properties will enable it to better understand human life, and assist with daily housework such as laundry sorting, clothes maintenance and organizing, or choosing clothes.

For perceiving material properties, tactile sensing is important.
\citet{Klatzky1993, HumanTactileTiest} demonstrated that humans use different exploratory procedures to sense different properties of objects, such as roughness or compliance. 
Researchers have been trying to make a robot to learn the material properties through touch as well. ~\citet{Katherine2013, xu2013tactile} developed setups to perceive properties of general objects using tactile sensors and a set of pre-set procedures, like squeezing and sliding. 
However, making robots explore the refined object properties in the natural environment remains a big challenge, and discriminating the subtle difference between the objects in the same category, such as clothing, is more difficult.
The challenge comes from two major sides: how to obtain adequate information from a tactile sensor, and how to generate an effective exploration procedure to obtain the information. 

\begin{figure}[t]
	\centering{
        \includegraphics[scale=0.98]{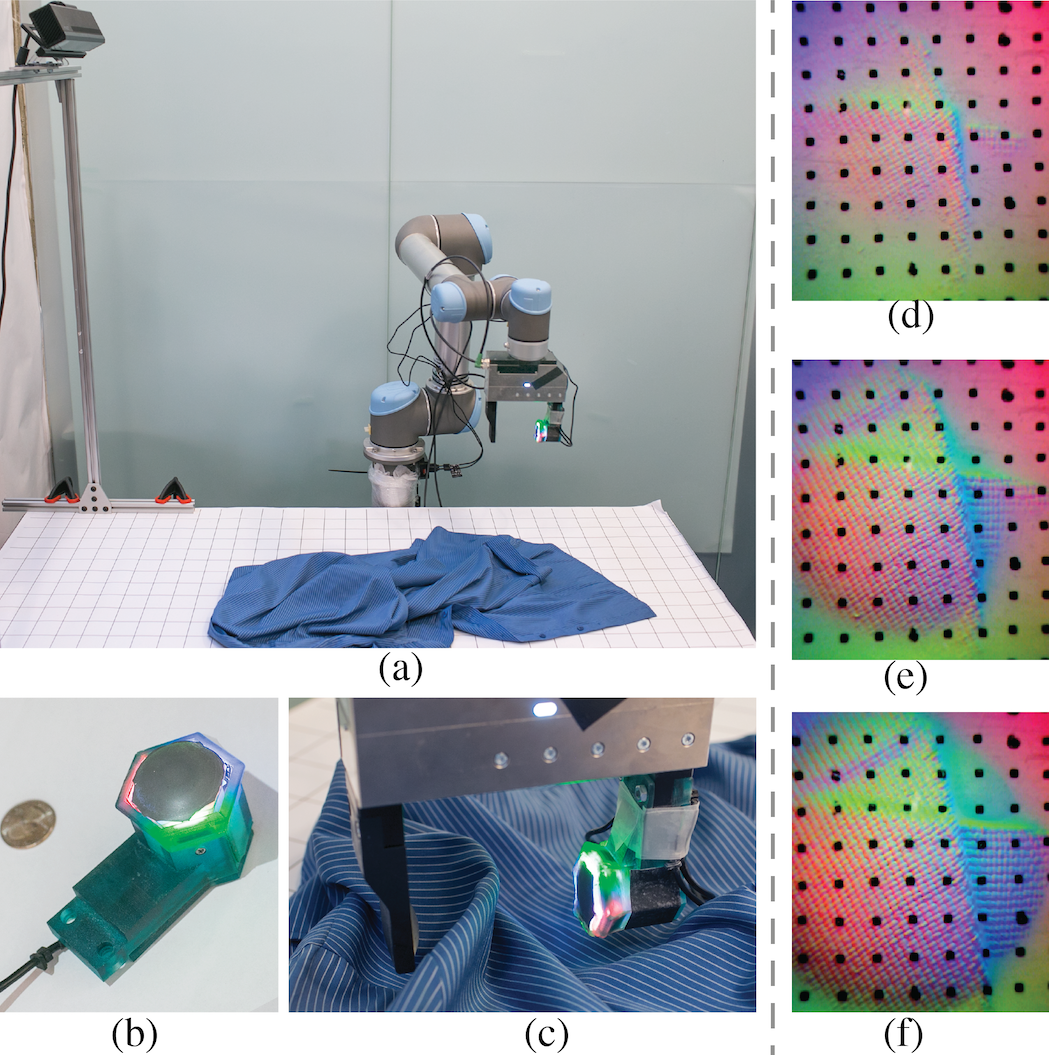}	
	}
	\caption{
    (a)The robotic system that automatically perceives clothes when they are in the natural environment. The system includes a robot arm, a gripper, a GelSight sensor mounted on the gripper, and a Kinect sensor. (b)The Fingertip GelSight sensor. (c)The gripper with a GelSight sensor mounted is gripping on the clothes.  (d)-(f) The tactile images from GelSight when gripping with increasing force. 
    }
	\label{fig:IntroFig}
\end{figure}
\begin{figure*}
	\centering{
    \includegraphics[scale=0.99]{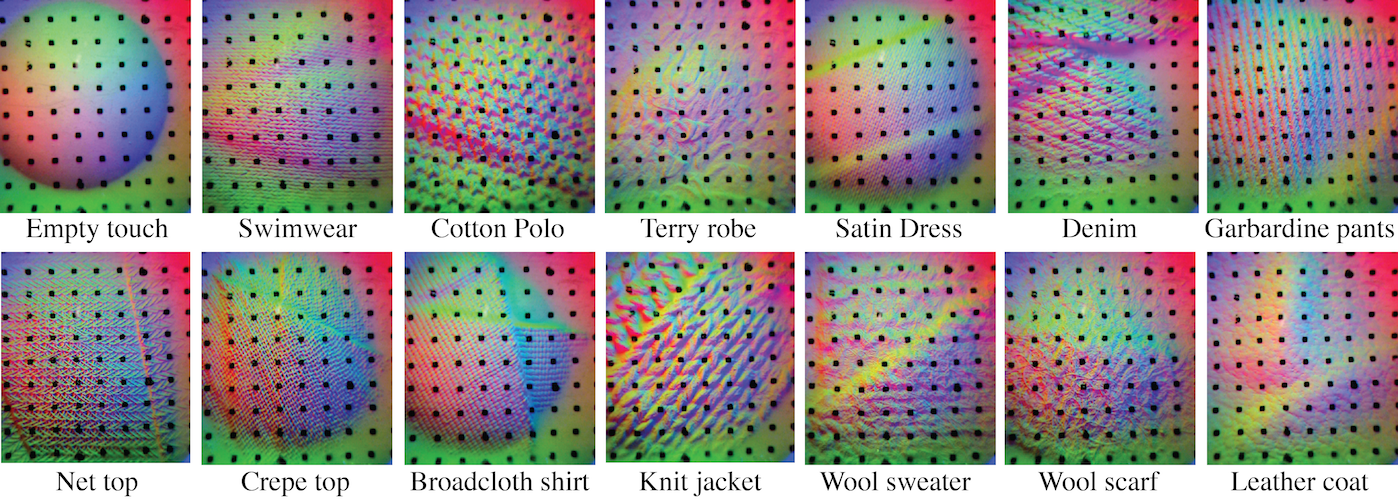}			
	}
	\caption{Examples of GelSight images when the robot squeezes on clothes(color rescaled for display purpose). Different clothes make different textures on GelSight images, as well as different overall shapes and folding shapes. The top left example shows the example when there is no clothing in the gripper.}
	\label{fig:GelSightSample}
\end{figure*}

At the same time, clothing related tasks have been a research interest for a long time, and the major focus has been in both the manipulation and recognition sides.
Most of the related works use only vision as sensory input, which measures the clothes' global shapes. Therefore, the clothing recognition is mostly restricted to the rough classification of the clothing type. The perception of fine-grained clothing properties, or the study on common clothes with a wide variety, is still undeveloped. 

In this paper, we design a robotic system that perceives the material properties of common clothes using autonomous tactile exploration procedures. The hardware setup of the system is shown in Figure~\ref{fig:IntroFig}(a). 
We address the two challenges of tactile exploration of object properties: to collect and interpret the high-resolution tactile data, and to generate exploration procedures for data collection.
The tactile sensor we apply is a GelSight sensor~\cite{GelSight2009,GelSight2011}, which senses the high-resolution geometry and texture of the contact surface. A GelSight sensor uses a piece of soft elastomer as the contact medium, and an embedded camera to capture the deformation of the elastomer. The exploration procedure is squeezing a part of the clothes, mostly a wrinkle, and recording a set of tactile images with GelSight (see Figure~\ref{fig:IntroFig}(c)-(f)). Then we train a Convolutional Neural Network (CNN) for multi-label classification to recognize the clothing properties. 
For generating exploration procedures autonomously, we use an external Kinect sensor to get the overall shapes of the clothes, especially the positions of the wrinkles, and train another CNN to pick up preferable points on the wrinkles. The robot will follow the Kinect detection for effective exploration. 
We also make the exploration closed-loop: if the tactile data is not good, which means the neural network cannot recognize the properties with high confidence, then the robot will re-explore the clothing on another location, until it gets good tactile data and confident results.

The 11 properties we studied are the physical properties, including thickness, fuzziness, smoothness, softness, etc., and the semantic properties that are more related to the application of the clothes, including wearing seasons, preferred washing methods, and textile type. The semantic properties could help robots to sort the clothes for multiple house chore tasks.
To make the system robust to a wide range of common clothes, we collect a dataset of 153 pieces of clothes for the training, and the dataset covers different clothing types, materials and sizes. 
Experimental results show that the system can recognize the clothing properties for both seen and unseen items, as well as detecting effective locations to generate tactile exploration. The robot can use the trained networks to do closed-loop exploration on the clothes. 
To our knowledge, this is the first work on studying fine-grained clothing properties with robot tactile sensing. 
The methodologies of this work will enable robots to understand common clothes better, and assist humans on more housework such washing laundry and clothing sorting.

\begin{figure*}
	\centering{
		\includegraphics[scale=1]{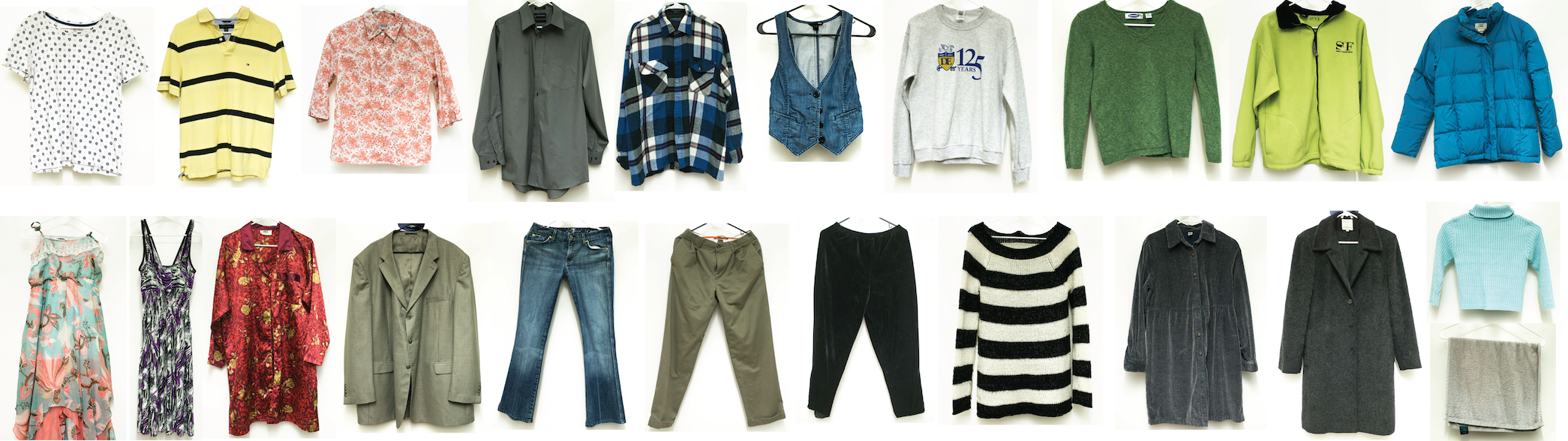}	
	}
	\caption{Examples of the clothes in the dataset. Our dataset contains 153 items of clothes in total, which range widely in different materials, sizes, and types. }
	\label{fig:ClothesPhotos}
\end{figure*}

\section{Related Work}
\subsection{Clothes classification}

The robotic community has been interested in clothing related topics for years, especially for the home assistant robotic tasks. The major focus has been clothing manipulation and recognition/classification. 
Researches on clothing manipulation are mostly about grasping, folding and un-folding. 
On the clothing recognition or classification tasks, most of the researches use vision as sensory input, and classify the clothes according to their rough types, such as pants, t-shirts, coats, etc. \citet{ClothClassification2011, LiRecognizing2014, CNNClassification2016} introduced methods for clothing classification by matching the 2D or 3D shape of the clothing to the clothing dataset. 
\citet{LiSunRecognition2016} proposed a method to recognize clothing type from stereo vision, where they applied more local features, such as the clothing's wrinkle shapes and textures. 

On multi-modal clothing perception, \citet{ClothesPhotometric2016} proposed a robotic system to classify clothes' general types and materials. They used an RGBD camera to capture the global shape, a photometric stereo sensor to record surface texture, and a fingertip tactile sensor to measure the dynamic force when rubbing the clothing. They showed that the multi-modal input, especially the texture perception from the photometric stereo sensors, largely improve the precision of the material recognition. 
However, recognizing fine-grained clothing properties of common clothes remains a challenge.

\subsection{Tactile sensors and GelSight}

Tactile sensing is an important sensory modality for robots. In the past decades, different kinds of tactile sensors have been developed, as reviewed in ~\cite{TactileReview2010, TactileManipulation2015}. The majority of the existing tactile sensors measure force or contact distribution over an area. 
Tactile sensing has been used for object shape classification(e.g. ~\cite{pezzementi2011tactile}), and for estimating material properties by combining with motion. \citet{Katherine2013} and \citet{xu2013tactile} show two systems of using tactile sensors to estimate multiple material properties. They used either a robot or a motion system to make the tactile sensor to press or slide on the object surface, and thus classifying different properties or adjective descriptions from the tactile signals.

In this work, we apply a GelSight tactile sensor~\cite{GelSight2009,GelSight2011}. The GelSight sensor is an optical-based tactile sensor that measures the surface geometry with very high spatial resolution (around 30 micros). Moreover, the printed markers on the sensor surface enable it to measure the contact force or shear~\cite{GelSightShear}. 
~\citet{GelSightTexture} showed that the high resolution of GelSight makes it very effective to discriminate different material categories by surface texture. The sensor can also estimate the physical properties of the objects through contact. An example is ~\cite{GelSightHardness17}, where the researchers showed that the GelSight signal can be used to estimate objects' hardness using the change of the contact shape under the increasing force. 
~\citet{GelSightFabrics} studied the GelSight's performance on fabric perception, where they tried to discriminate different fabrics using an embedding vector, which describes the overall properties of the fabrics, from either the vision or tactile input.

\subsection{Deep learning for texture recognition and tactile perception}
Texture provides significant information about material properties. Early works on vision-based texture recognition mostly used hand-crafted features like Textons~\cite{julesz1981textons}, Filter Banks~\cite{de1997multiresolution} or Local Binary Patterns(LBP)~\cite{ojala2002multiresolution}.
In recent years, The Convolutional Neural Networks (CNN) have achieved many state-of-the-art performances on computer vision tasks
and were successfully applied to texture recognition: ~\citet{cimpoi2015deep} proposed FV-CNN, which combined the CNN with Fisher Vectors(FV) to better extract localized features. The convolutional layers in FV-CNN are from the VGG model~\cite{simonyan2014very}, pre-trained on ImageNet~\cite{deng2009imagenet}, and served as filter banks; the FV was used to build the orderless representation. ~\citet{andrearczyk2016using} proposed T-CNN for texture classification, which used an energy layer after the convolutional layers.

The CNN models developed for computer vision also proved effective for processing tactile information: ~\cite{GelSightHardness17} and ~\cite{GelSightFabrics} used the CNNs on GelSight data images for estimating object hardness or fabric properties, while the networks are pre-trained on normal images. 

\section{Data Collection}

The aim of this project is to develop a robotic system that can autonomously perceive clothes and classify them according to material properties. 
The robot's hardware system consists of a robot arm and a gripper with the tactile sensor GelSight. We use an external Kinect camera to guide the robot exploration, but only for planning the motion. The robot explores the clothing by squeezing a part of it, and during the process, the GelSight sensor will record a sequence of tactile images of the clothing' local shape. 
We collect 2 kinds of data: the GelSight image sequences, and the gripping points on the Kinect depth images.
The GelSight images help the robot to recognize the clothing properties, and the depth images and the exploration results help the robot to learn whether a gripping position is likely to generate good tactile data.

\subsection{Clothing Dataset}
We collect a dataset of 153 pieces of new and second-hand clothes. The clothes are all common clothes in everyday life, but widely distributed on types, materials and sizes. We aim to make the dataset cover all kinds of common clothes in an ordinary family. The dataset also includes a small amount of other fabric products, such as scarfs, handkerchiefs and towels. Some examples of the clothes are shown in Figure~\ref{fig:ClothesPhotos}.


\begin{table}
\begin{center}
\caption{Clothing property labels}
\label{tab:clothinglabel}
	\begin{tabular}{|p{3.8cm}|p{3.8cm}| }
    \hline
    Thickness (5) & Smoothness (5)\\
    \hline
    0 - very thin (\textit{crepe dress}) &  0 - very smooth (\textit{satin})\\
    1 - thin (\textit{T-shirt})       & 1 - smooth (\textit{dress shirt})  \\
    2 - thick(\textit{sweater})   & 2 - normal (\textit{sweater})\\
    3 - very thick (\textit{woolen coat})  & 3 - not smooth (\textit{fleece}) \\
    4 - extra thick (\textit{down coat})  & 4 - rough (\textit{woven polo}) \\
    \hline
    \hline
    Fuzziness (4)  &  Season (4) \\
    \hline
    0 - not fuzzy (\textit{dress shirt})  & 0 - all season (\textit{satin pajama}) \\
    1 - a little fuzzy (\textit{dress shirt})  & 1 - summer (\textit{crepe top}) \\
    2 - a lot fuzzy (\textit{terry robe})  & 2 - spring/fall (\textit{denim pants}) \\
    & 3 - winter (\textit{cable sweater}) \\
    \hline
    \hline
    Textile type (20) & Washing method (6) \\
    \hline
    \parbox[t]{3.8cm}{cotton; satin; polyester; denim; garbardine; broad cloth; parka; leather; crepe; corduroy; velvet; flannel; fleece; hairy; wool; knit; net; suit; woven; other}    
    &
    \parbox[t]{3.8cm}{machine wash warm; machine wash cold; machine wash cold with gentle cycles; machine wash cold, gentle cycles, no tumble dry; hand wash; dry clean}
    \\
    \hline
    \hline
    \multicolumn{2}{|l|}
    {\parbox[t]{7.6cm}{Labels with binary classes: \\
    Softness, stretchiness, durability, woolen, wind-proof 
        }}
\\     \hline
    \end{tabular}    
\end{center}
\end{table}

The property labels we choose are a set of common properties that humans use to describe clothes. We used the 11 labels, with either binary classes or multiple classes. The labels and examples of the classes are shown in Table~\ref{tab:clothinglabel}.



\subsection{Robotic System Setup}

The robotic hardware system is shown in Figure~\ref{fig:IntroFig}(a), and it consists of 4 components: a robot arm, a robot gripper, a GelSight tactile sensor, and a RGBD camera. 
The arm is a 6 DOF UR5 collaborative robot arm from Universal Robotics, with a reach radius of 850mm and payload of 5kg. 
The parallel robotic gripper is a WSG 50 gripper from Weiss Robotics, with a stroke of 110mm, and a rough force reading from the current. We mount GelSight on the gripper as 1 finger, and the other finger is 3D printed with a curved surface, which helps GelSight get in full contact with the clothes. 
The GelSight sensor we used is the revised Fingertip GelSight sensor~\cite{GelSightNewsensor}, as shown in Figure~\ref{fig:IntroFig}(b). 
The sensor has a soft and dome-shaped surface for sensing, and a sensing range of 18.6mm$\times$14.0mm, spatial resolution of 30 microns for geometry sensing. The elastomer on the sensor surface is about 5 Shore A scale, and the peak thickness is about 2.5mm. The sensor collects data at a frequency of 30Hz. 
The external RGBD camera we used is a Kinect 2 sensor, which has been calibrated and connects to ROS system via IAI Kinect2~\citep{iai_kinect2} toolkit. It is mounted on a fixed supporting frame which is 106mm above the working table and a tilt angle of \ang{23.5}, so that the sensor is able to capture a tilted top view of the clothes.

Note that in the grasping procedure, due to the repetitive large shear force, the GelSight surface would wear off after a series of grasping, so that we have to change the sensing elastomer for multiple times. Since the elastomer sensors are manually made, they have slightly different marker patterns and shapes, which result in some differences in the tactile images.

\subsection{Autonomous Data Collection}

\begin{figure}
	\centering{
		\includegraphics[scale=1]{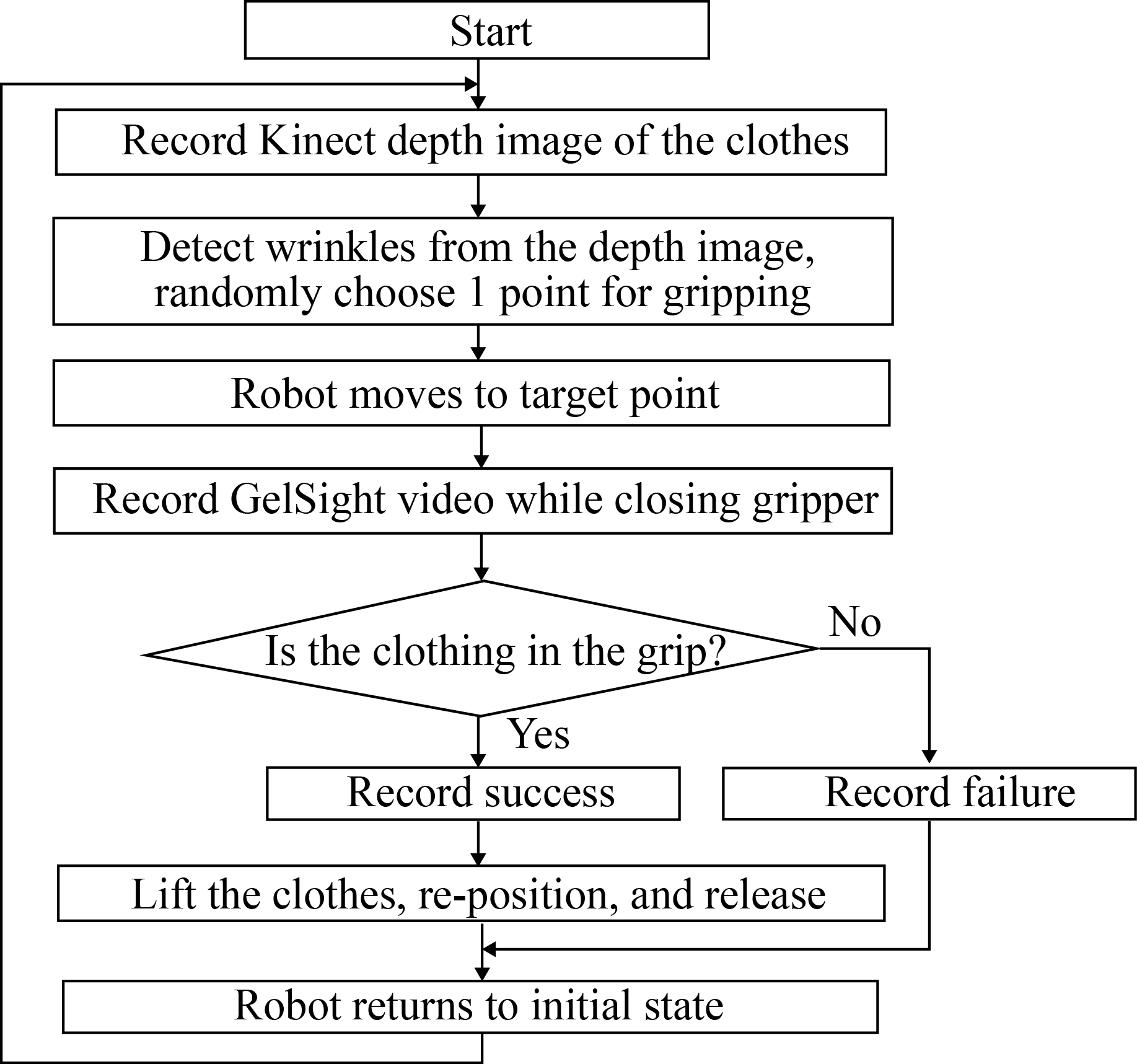}				
	}
	\caption{The flow chart of the autonomous data collection process. }
	\label{fig:CollectFlow}
\end{figure}

The training data is autonomously collected by the robot. The flow chart of the process is shown in Figure~\ref{fig:CollectFlow}. 


\begin{figure}
	\centering{
		\includegraphics[scale=1]{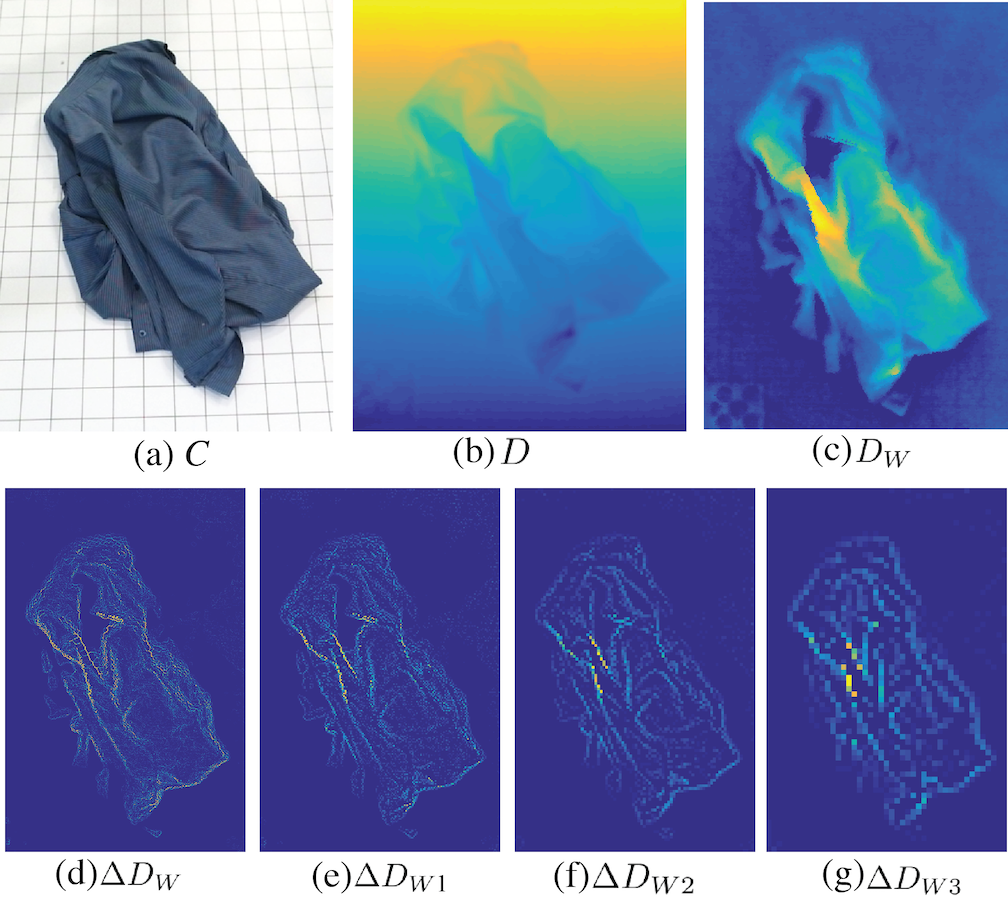}				
	}
	\caption{(a) The RGB image from Kinect. (b) The depth image $D$ from Kinect. (c) $D_W$: the depth image in the world coordinate. (d) $\Delta D_W$: the Laplacian operated $D_W$, where the borders are picked. (e)-(g): the Laplacian operated $D_W$ on different pyramid levels. The color of the figures are re-scaled for display purposes.}
	\label{fig:KinectView}
\end{figure}

\textbf{Choosing gripping positions from Kinect images}
The robot is most likely to collect good tactile data when it grips on the wrinkles on the clothes. The wrinkles are higher than the surrounding area, which will be captured by Kinect's depth images $D$. We firstly transfer the depth map into the world frame, thus obtain the depth map $D_W$ using 
\begin{equation}
	D_W=T_{K2W} \cdot K^{-1}\cdot D
\end{equation}
where $K$ is the camera matrix that expanded to $4\times4$ dimension, and $T_{K2W}$ is the $4\times4$ transformation matrix from the Kinect frame to world frame. 
We set the $x-y$ plane in the world frame as the table, so that the `depth' value of $D_W$, which is represented as $z$, corresponds to the real height of the clothes on the table. An example of the transformed $D_W$ is shown in Figure~\ref{fig:KinectView}(c).
The edges of $D_W$, which could be easily picked by Laplacian operation, show the wrinkles on the clothes. We apply the pyramid method to down-sample the image to 3 different levels, therefore the high-derivative areas on different levels represent the wrinkles of different widths. 
From all the high-derivative points in the 3 levels, we randomly choose 1 point as the target gripping position.

Before gripping, the gripper should rotate to the angle perpendicular to the wrinkle. We calculate the planar direction of the wrinkle at point $(x,y)$ by 
\begin{equation}
   Dir(x,y)=\arctan (\frac{\partial D_W(x,y)}{\partial y} / \frac{\partial D_W(x,y)}{\partial x}) 
\end{equation}

\textbf{Gripping on the wrinkles}
Once the target point on the wrinkle is selected, the robot will move about the point, with the gripper in a perpendicular direction, and then descend to the position below the wrinkle to grip the clothing, with a low speed of 5mm/s. 
The gripper stops closure when the motor current reaches a threshold, which indicates a large impedance force. The GelSight records videos during the closure. Typically the GelSight records 10 to 25 frames for one gripping iteration.

After the gripping, we judge whether the contact is valid using GelSight images. If the GelSight image shows no contact with the clothing, we mark this tactile data invalid, and mark the gripping location as a failure case. 

\section{Clothes Classification using Deep Learning}
\label{chpt:network}

\begin{figure*}
	\centering{
		\includegraphics[scale=1]{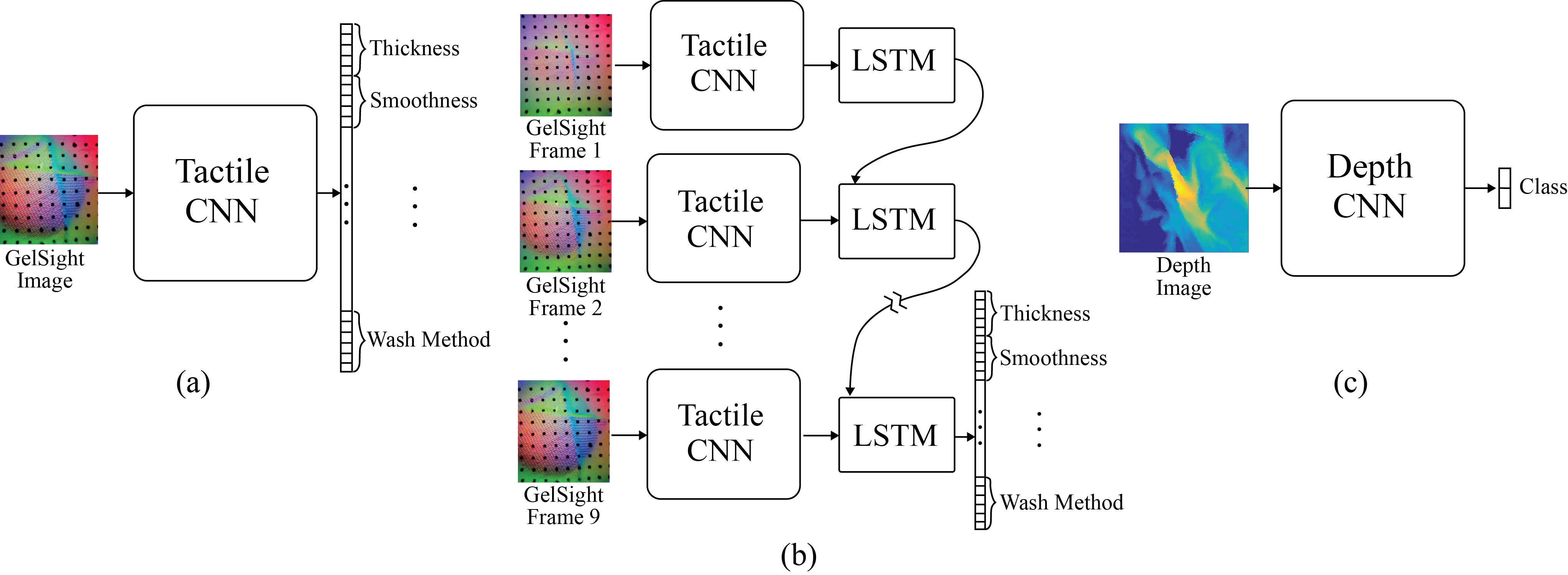}				
	}
	\caption{(a) The multi-label classification network for recognizing different properties from a single GelSight image. (b) The neural network for recognizing different properties from GelSight video, where we choose 9 frames from the video as the input. (c) The network for evaluating whether the gripping point would generate effective tactile data. For the Tactile CNN, we apply VGG19; for the Depth CNN, we apply VGG16. }
	\label{fig:Network}
\end{figure*}

In this project, we designed two separate neural networks for 2 independent goals: 1) selecting a point on the clothing for the robot to explore, and 2) estimating the properties of the clothing from the collected tactile data. 

\subsection{Networks for property perception}
To perceive the properties of the clothes, we use a CNN for the multi-label classification of the GelSight images. The labels correspond to the clothing properties, and they are independently trained. We use two kinds of networks: one takes a single GelSight image as the input (Figure~\ref{fig:Network}(a)), and the other one takes in multiple frames (Figure~\ref{fig:Network}(b)). 
The CNN for GelSight images are VGG19~\cite{VGG}, which is originally designed for object recognition for general images, and pre-trained on the image dataset ImageNet~\cite{deng2009imagenet}.

For the network with a single input frame, we choose the GelSight image when the contact force is the maximum, and use a single CNN to classify the image. For recognizing the multiple properties, we train the same CNN with classification on multiple labels, which correspond to the clothing properties. The architecture is shown in Figure~\ref{fig:Network}(a). 

Additional to learning the properties from the single GelSight image, we also try to learn the properties from the GelSight image sequence. The sequence includes a set of images when the sensor squeezes the clothing with increasing forces, thus the frames record the surface shapes and textures under different forces. The image sequences are more informative than the single images. To train on the image sequence, we use the structure connecting CNN and a long short-term memory units (LSTM)~\cite{LSTM} with a hidden state of 2048 dimensions, as shown in Figure~\ref{fig:Network}(b). We use the features from the second last layer fc6 from VGG16 as the input of LSTM.

The image sequence contains 9 frames, with a equal time stamp interval until reaching the frame of max contact. We choose the number of 9 as a balance of low computational cost and the sum of information. 
Since the gripper closes slowly and evenly when collecting the data, the gripper's opening width between the frames is equal. As a result, some of the thick clothes would deform largely in the squeezing process, so that the selected sequence starts after the contact; while when gripping thin clothes, the maximum contact point is easily reached, and the selected sequence starts with several blank images. 

\subsection{Networks for gripping point selection}
We train a CNN (based on VGG16~\cite{VGG} architecture) to learn whether the gripping location is likely to generate good tactile data. The network architecture is shown in Figure~\ref{fig:Network}(c). The input data is a cropped version of $D_W$, the depth image in the world frame, and the output is a binary class on whether the image represents a potentially successful gripping. 
To indicate the gripping location in the depth image, we crop the depth image to make the gripping location the center of the new image, and the window size is 11cm$\times$11cm. 

\subsection{Offline training of the neural networks}

We divided the perception data from the 153 items of the clothes into 3 sets: the training set, the validation set, and the test set. The training set and validation set make of data from 123 items of clothes, and the testing set contains data from the rest 30 items. For the 123 items, we randomly choose data from 85\% of collecting iterations as the training set, and 15\% of the data as the validation set.
The division of clothes for training and testing is manually done ahead of the network training, with the standard that the clothes in the test set should be a comprehensive representation of the entire dataset.

In all the exploration iterations, we consider 2 situations that the exploration is `failed':1) the gripper does not contact the clothing, which can be detected automatically from GelSight data. 2) the contact is not good, that collected GelSight images is not clear. Those cases are manually labeled. We train the tactile CNNs with only the data from `successful' exploration. When training the Depth CNN, the iteration that are considered `successful' is made class 1.


We train the networks using stochastic gradient descent as the optimizer.
The weights of the Depth CNN (Figure~\ref{fig:Network}(c)) and single GelSight image(Figure~\ref{fig:Network}(a)) is pre-trained on ImageNet~\cite{deng2009imagenet}, and the CNN for the multi GelSight image input (Figure~\ref{fig:Network}(b)) is initialized with the weights of Figure~\ref{fig:Network}(a). For the video network, we jointly train the CNN and LSTM for 500 epocs, at a dropout rate of 0.5.


For training the network for GelSight images, we apply data augmentation to improve the performance of the network, by adding random values to the image intensity in the training. When training with the image sequence, we choose the input sequence slightly differently on the time stamp.

\subsection{Online robot test with re-trials}
We run the robot experiment online with the two networks: at the start of the exploration, the robot generates a set of candidate exploration locations from the depth image, and use the depth CNN to select a best one.
After collecting tactile data by gripping the clothing at the selected location, we use the tactile CNN to estimate the clothing properties. At the same time, the robot evaluates whether the collected tactile data is good, by analyzing the output classification probability of the tactile network. If the probability is low, it is likely the tactile data is ambiguous and the CNN is not confident about the result. In this case, the robot will explore the clothing again, until a good data point is collected. In the experiment (Section~\ref{chpt:onlinerobot}) we choose the property of washing method and the probability threshold of 0.75.

\section{Experiment}
We conduct both offline and online experiments. For the offline experiments, We use the data that the robot collected with 6616 iterations, which includes 3762 valid GelSight videos, while the rest 2854 iterations did not generate good data because of inadequate gripping locations. The invalid data, is picked either autonomously by GelSight or manually.
The dataset is available at \url{http://data.csail.mit.edu/active_clothing/Data_ICRA18.tar}.

\subsection{Property perception}
\begin{table}
\begin{center}
\caption{Result of property perception on seen and unseen clothes}
\label{tab:GelSight_result}
	\begin{tabular}{|c|c|c|c|c|c| }
    \hline
     & &\multicolumn{2}{c|}{Seen clothes} & \multicolumn{2}{c|}{Unseen clothes} \\
    \cline{3-6}
    &Chance & Image & Video & Image & Video\\
    \hline
    \hline
    Thickness &  0.2 & 0.89 &   0.90  & 0.67 &  0.69 
    \\ \hline
    Smoothness & 0.2 & 0.92 &  0.93  & 0.76 &  0.77 
    \\ \hline
    Fuzziness & 0.25 & 0.96 &  0.96 & 0.76 &  0.76 
    \\ \hline
    Softness & 0.5  & 0.95 &  0.95  & 0.72 &  0.76 
    \\ \hline
    Stretchiness & 0.5 & 0.98 & 0.98 & 0.80 &  0.81 
    \\ \hline
    Durability & 0.5 & 0.97 &  0.98   & 0.95 &  0.97
    \\ \hline
    Woolen & 0.5 & 0.98 &  0.98  & 0.90 &  0.89 
    \\ \hline
    Wind-proof & 0.5 & 0.96 &  0.96 & 0.87 &  0.89 
    \\ \hline
    Season & 0.25 & 0.89 &  0.90 & 0.61 & 0.63
    \\ \hline
    Textile type & 0.05 & 0.85 &  0.89 & 0.44 &  0.48 
    \\ \hline
    Wash method & 0.17 & 0.87 & 0.92  & 0.53 &  0.56 
    \\ \hline

    \end{tabular}    
\end{center}
\end{table}

In the experiment of property perception, we use 3762 GelSight videos from the 153 clothing items,
and classify the tactile images according to the 11 property labels. The training set includes 2607 videos, the validation set includes 400 videos from the same clothes, and the test set includes 742 videos from novel clothes. We try the networks with either a single image as input, or multiple images from a video as input. 
The results are shown in Table~\ref{tab:GelSight_result} 

From the results, we can see that for both seen and novel clothes, the networks can predict the properties with a precision much better than chance. Specifically, the precision on seen clothes is very high. 
However, the precision gap between the validation set and test set indicates the model overfits to the training set. We suppose the major reasons for the overfit are:
\begin{itemize}
\item The dataset size is limited. Although the dataset has a wide variety of clothing types, the number of the clothes in each refined category is small (2 to 5). 
\item We used 5 GelSight sensors in data collection, and they have some different optical properties, which result in some difference in the images. 
\item The CNNs are designed for visual images, which is not the optimum for the GelSight images. 
\item Some properties, are not only related to the materials but also the occasions of the clothing. For example, satin is mostly used for summer clothes, but a satin pajama, which feels exactly the same, is worn for all seasons.
\end{itemize}

We also experiment with other CNN architectures for the multi-label classification task, including VGG16 and AlexNet~\cite{krizhevsky2012imagenet}, but the results are not satisfactory. VGG19 performs relatively better. We suppose for the given task of tactile image classification, AlexNet and VGG16 are not deep enough to extract all the useful features.

Unfortunately, the neural network trained on videos (Figure~\ref{fig:Network}(b)) does not make a significant improvement. The possible reason is that the networks are overfit on the textures of the clothing, and the training set is not large enough to train the neural networks to learn the information from the dynamic change of the GelSight images.

We believe given enough resource, that we can collect a much larger clothing dataset, the property perception with unseen clothes will significantly improve. Another possible improving direction is to explore network structures that are more suitable to the tactile images, instead of the ones developed for general images. 

 \subsection{Exploring planning}
 

We experiment on picking effective gripping locations from the Kinect depth image, using the Kinect images from the 6616 exploration iterations. The images are also divided into the training set, validation set (on the same clothes), and test set (on unseen clothes). 
On both the validation and test sets, the output of the neural network has a success rate of 0.73 (chance is 0.5).
The result indicates the identification of the clothing item has limited influence on the result of gripping location selection. In the training process, the network quickly reaches the point of best performance and starts to overfit. For achieving better results for exploration planning, we plan to develop a more robust grasping system, and collect more data or use online training. 

\subsection{Online robotic test}
\label{chpt:onlinerobot}

In this experiment, the robot runs the exploration autonomously using the depth CNN and tactile CNN.
In the exploration, if the property estimation from the tactile CNN is not confident, which is most likely caused by the bad tactile data, the robot will re-do the exploration.
The tactile CNN in this experiment is the CNN for single image input (Figure~\ref{fig:Network}(a)). 

We experiment on the test clothes(30 items), and each clothes is explored 5 times.
The result is shown in Table~\ref{tab:GelSight_result_online}. Here we compared the result of `without re-trial' which means the system would not judge the data quality, and `with re-trial'. Note that the `without re-trial' results are worse than the results in Table~\ref{tab:GelSight_result} because the tactile data here is all the raw data generated by the robot, while Table~\ref{tab:GelSight_result} is only from good data. Another reason is that the gel sensor in this experiment is a different one, and not seen in the training set before, so that there is some slight difference in the lighting distribution. 
The results also showed that with the re-trials, the precision of property classification increases largely. On average, the robot makes 1.71 trials for each exploration, but 77.42\% of the clothes are `easy' for the robot, that it takes less than 2 grasps to get a confident result, and it turns out the property estimation is more precise. The rest clothes are more `confusing', that the robot need to explore them for multiple times, but the properties are still not well recognized.

\begin{table}
\caption{Property perception on unseen clothes in online robot test}
\label{tab:GelSight_result_online}
\begin{center}
\begin{tabular}{|c|c|c|c|c|}
\hline
 & & Without &  With & With Re-trial,
 \\ 
 Properity & Chance & Re-trial & Re-trial & on Easy Clothes \\ \hline
\hline
Thickness      &0.2     & 0.59 & 0.65 & 0.72\\ \hline
Smoothness     &0.2 	& 0.71 & 0.74 &  0.82\\\hline
Fuzziness      &0.25     & 0.67 & 0.74 &0.82\\ \hline
Softness       &0.5     & 0.60 & 0.66 & 0.72\\\hline
Stretchiness   &0.5		& 0.74 & 0.81 & 0.88\\\hline
Durability      &0.5	 	& 0.86 & 0.86 &0.91 \\\hline
Woolen  &0.5        & 0.92 & 0.91 &0.93\\\hline
Wind-proof 	&0.5	& 0.83 & 0.82 & 0.86\\\hline
Season         &0.25 & 0.57 & 0.64 & 0.71\\\hline
Textile type   &0.05      & 0.37 & 0.50  & 0.59\\\hline
Wash Method  	&0.17	& 0.50 & 0.60 & 0.71\\\hline

\hline
\end{tabular}
\end{center}
\end{table}

\section{Conclusion}

The perception of object properties is an essential part to make an intelligent robot to understand and interact with the environment, and among the common objects, clothing is an important category. By better understanding clothing properties, the robots will be able to better understand the humans' lives, and better assist humans in housework. 
In this paper, we introduce a robotic system that autonomously perceives the material properties of the common items of clothes. The system uses a GelSight tactile sensor as the sensory input, and recognize 11 properties of the clothes, which helps the robot to have a relatively comprehensive understanding of the clothing material. We use Convolutional Neural Networks (CNN) to classify the GelSight tactile images, and the model can well recognize clothing properties of seen clothes, and effectively recognize the unseen clothes. 
At the same time, we use a Kinect sensor to guide the robot to explore the clothes in a natural environment, and use a method that combines a hand-crafted model and CNN to find the effective contact locations. 
To our knowledge, this is the first work on perceiving fine-grained properties of common clothes, and provides a new example of active tactile perception of object properties for robots. The system and methodologies in this paper will help robots to better assist humans in clothing related housework, and inspiring other works on active tactile sensing on object properties. 

\addtolength{\textheight}{0cm}


\section*{ACKNOWLEDGMENT}
The work is financially supported by Toyota Research Institute. The authors sincerely thank John-Elmer Canfield and Christopher Cullity for the help on building the clothes dataset and labeling the properties, thank Siyuan Dong for help building the GelSight sensor, and thank MIT Lincoln Lab and NTT Communication Sciences Laboratories.


\bibliographystyle{IEEEtranN}
\bibliography{Ref_Clothes}

\end{document}